# Deep Octonion Networks

Jiasong Wu[1, 2, 3, 4], Ling Xu[1, 4], Youyong Kong[1, 4], Lotfi Senhadji[2, 3, 4], Huazhong Shu[1,4]

[1]*Laboratory of Image Science and Technology, Key Laboratory of Computer Network and Information Integration, Southeast University, Ministry of Education, Nanjing 210096, China*

[2]*INSERM, U1099, Rennes, F-35000, France*

[3]*Université de Rennes 1, LTSI, Rennes, F-35042, France*

[4]*Centre de Recherche en Information Médicale Sino-français (CRIBs)*

**Abstract**: Deep learning is a research hot topic in the field of machine learning. Real-value neural networks (Real NNs), especially deep real networks (DRNs), have been widely used in many research fields. In recent years, the deep complex networks (DCNs) and the deep quaternion networks (DQNs) have attracted more and more attentions. The octonion algebra, which is an extension of complex algebra and quaternion algebra, can provide more efficient and compact expression. This paper constructs a general framework of deep octonion networks (DONs) and provides the main building blocks of DONs such as octonion convolution, octonion batch normalization and octonion weight initialization; DONs are then used in image classification tasks for CIFAR-10 and CIFAR-100 data sets. Compared with the DRNs, the DCNs, and the DQNs, the proposed DONs have better convergence and higher classification accuracy. The success of DONs is also explained by multi-task learning.

**Keywords:** Convolutional neural network, complex, quaternion, octonion, image classification

## 1. Introduction

Real-value neural networks (Real NNs) [1-12] attracted the attention of many researchers and recently made major breakthroughs in many areas such as signal processing, image processing, natural language processing, etc. Many models of Real NNs have been constructed in the literature. These models can generally be categorized into two kinds: non-deep models and deep models. The non-deep models are mainly constructed by multilayer perceptron module [13] and hard to train, if we only use the real-valued back propagation (BP) algorithm [14], when their layers are larger than 4. The deep models can be roughly constructed by the following two strategies: multilayer perceptron models assisted by the unsupervised pretrained methods (for example, deep belief nets [15], deep auto-encoder [16], etc.) and real-value convolutional neural networks (Real CNNs), including LeNet-5 [17], AlexNet [18], Inception [19-22], VGGNet [23], HighwayNet [24], ResNet [25], ResNeXt [26], DenseNet [27], FractalNet [28], PolyNet [29], SENet [30], CliqueNet [31], BinaryNet [32], SqueezeNet [33], MobileNet [34], etc.

Although Real CNNs have achieved great success in various applications, the correlations between convolution kernels generally do not take into consideration, that is, there are no connections or no special relationships considering between convolution kernels. The opposite of Real CNNs is real-value recurrent neural networks (Real RNNs) [35-38], who obtain the correlations by adding the connections between convolution kernels and then learn the weights of these connections, which, however, increased significantly the training difficulty and was easier to encounter converge problems. *The first question has been raised: Can we consider the correlations between convolution kernels by some special relationships, which do not need to learn, instead of adding the connections between convolution kernels?*

Many researchers find that the performance can be improved when the relationships between convolution kernels are modeled by complex algebra, quaternion algebra [39-42], and also octonion algebra [43-45]. Therefore, it attached a lot of attention for extending the neural networks from real domain to complex domain, quaternion domain, and also octonion domain. These extension models, like Real CNNs, can also be divided into two categories: non-deep models [46-60] and deep models [61-71]. There are many research work focus on the non-deep models, for example, Widrow *et al*. [46] first introduced the complex-valued neural networks (Complex NNs), which have been widely used in recent years in radar imaging, image processing, communication signal processing, and many others [47]. Compared to the Real NNs, the Complex NNs have better generalization ability due to their time-varying delays and impulse effects [48]. Arena *et al*. [49] then extended the neural networks from complex domain to quaternion domain and proposed quaternion-valued neural networks (Quaternion NNs), which have been applied to color image compression [50], color night vision [51], and 3D wind forecasting [52]. Furthermore, the quaternion-valued BP algorithms achieve correct geometrical transformations in color space for an image compression problem, whereas real-valued BP algorithms fail [53, 54]. Popa [55] further extended the neural networks from quaternion domain to octonion domain and proposed octonion-valued neural networks (Octonion NNs), whose leakage delays, time-varying delays, and distributed delays were also introduced [56, 57]. Moreover, Clifford-valued neural networks [58-60] were also proposed for the extension of complex NNs and quaternion NNs. For the deep models, Reichert and Serre [61] proposed complex-valued deep networks, which interpreted half of the cell state as an imaginary part and used complex values to simulate the phase dependence of biologically sound neurons. Then, some researchers proposed complex-valued convolutional neural networks (Complex CNNs) [62-65]. Among them, Trabelsi *et al*. [65] proposed deep complex networks (DCNs) which provide the key atomic components (complex convolution, complex batch normalization, and complex weight initialization strategy, etc.) for the construction of DCNs and also obtain lower error rate than corresponding deep real networks (DRNs) [25] in CIFAR-10 and CIFAR-100 [66]. Then, Gaudet and Maida [67] extended the deep networks from complex domain to quaternion domain and proposed deep quaternion networks (DQNs), which achieve better image classification performance than DCNs [65] in CIFAR-10 and CIFAR-100. Meanwhile, Titouan *et al*. also proposed deep quaternion neural network (QDNN) [68, 69], quaternion recurrent neural network (QRNN) [70], bidirectional quaternion long-short term memory (BQLSTM) [71], and quaternion convolutional autoencoder (QCAE) [72]. *The second question has been raised: Can we extend the deep networks from quaternion domain to octonion domain to obtain a further benefit? How to explain the success these deep networks on these wider number domains (complex, quaternion, octonion)*?

In order to solve the second question, in this paper, we proposed deep octonion networks which can be seen as an extension of the deep networks from quaternion domain to octonion domain. The contributions of the paper are as follows:

1）The key atomic components of deep octonion networks are introduced and implemented, for example, octonion convolution module, octonion batch normalization module and octonion weight initialization method.

2）When applying the proposed deep octonion networks on the classification tasks on CIFAR-10 and CIFAR-100, the classification results are better than deep real networks, deep complex networks and deep quaternion networks.

3) The explanation of deep complex networks, deep quaternion networks, and deep octonion networks from the perspective of multi-task learning [73-75].

The rest of the paper is organized as follows. In Section 2, octonion representation, its main properties and characteristics are briefly introduced. The architectural components needed to build deep octonion networks is described in Section 3. The classification performance of deep octonion networks is analyzed and also compared to the deep real networks, deep complex networks, and deep quaternion networks in Section 4. Then, Section 5 explains deep networks on these wider number domains from the perspective of multi-tasking learning. The conclusions are formulated in Section 6.

## 2. Octonion Representation

An octonion number $x$ is a hypercomplex number, which is an extension of complex number and quaternion number, consists of one real part and seven imaginary parts:

$$x = x_0 e_0 + x_1 e_1 + x_2 e_2 + x_3 e_3 + x_4 e_4 + x_5 e_5 + x_6 e_6 + x_7 e_7 \in O \tag{1}$$

where $O$ denotes the octonion number field, $x_i \in R, i=1,2,...,7$ ($R$ denotes the real number field), $e_0=1$, and $e_i$, $i=1,2,...,7$, are seven imaginary units obeying the following rules:

$$\begin{cases} e_i^2 = -1 \\ e_i e_j = -e_j e_i \neq e_j e_i \\ (e_i e_j) e_k = -e_i (e_j e_k) \neq e_i (e_j e_k) \end{cases}, \quad \forall i \neq j \neq k, \ 1 \leq i, j, k \leq 7, \tag{2}$$

The above equation shows that the octonion multiplication is neither commutative nor associative. The multiplication tables of the imaginary units are also shown in Table 1.

The conjugate of this octonion $x \in O$ is given by

$$x^* = x_0 e_0 - x_1 e_1 - x_2 e_2 - x_3 e_3 - x_4 e_4 - x_5 e_5 - x_6 e_6 - x_7 e_7 \ . \tag{3}$$

The unit octonion of $x \in O$ is

$$x^{\triangleleft} = \frac{x}{\sqrt{x_0^2 + x_1^2 + x_2^2 + x_3^2 + x_4^2 + x_5^2 + x_6^2 + x_7^2}} \ . \tag{4}$$

For a complete review of the properties of octonion, please refer to [43].

## 3. Deep Octonion Networks

This section introduces the methods and the modules to construct the deep octonion networks, which include octonion internal representation method (Section 3.1), octonion convolution module (Section 3.2), octonion batch normalization module (Section 3.3), and octonion weight initialization method (Section 3.4).

### 3.1 Octonion internal representation method

The deep octonion networks are the octonion extension of the well-known deep real networks [25], deep complex networks [65] and deep quaternion networks [67]. We represent the real part and seven imaginary parts of an octonion number as logically distinct real valued entities and simulate octonion arithmetic using real-valued arithmetic internally. If we assume that an octonion convolutional layer has $N$ feature maps where $N$ is divisible by 8, then, these feature maps can be split into 8 parts to form an octonion representation. Specifically, as shown in Fig. 1, we allocate the first $N/8$ feature maps to the real part and the remaining seven $N/8$ feature maps to the seven imaginary parts.

### 3.2 Octonion convolution module

Octonion convolution can be implemented by convolving an octonion vector $\mathbf{x}_o \in O^N$ by an octonion filter matrix $\mathbf{W}_o \in O^{N \times N}$ as follows:

$$\begin{aligned}
\mathbf{W}_o *_o \mathbf{x}_o &= (\mathbf{W}_0 + \mathbf{W}_1 e_1 + \mathbf{W}_2 e_2 + \mathbf{W}_3 e_3 + \mathbf{W}_4 e_4 + \mathbf{W}_5 e_5 + \mathbf{W}_6 e_6 + \mathbf{W}_7 e_7) \\
&\quad *_o (\mathbf{x}_0 + \mathbf{x}_1 e_1 + \mathbf{x}_2 e_2 + \mathbf{x}_3 e_3 + \mathbf{x}_4 e_4 + \mathbf{x}_5 e_5 + \mathbf{x}_6 e_6 + \mathbf{x}_7 e_7) \\
&= (\mathbf{W}_0 * \mathbf{x}_0 - \mathbf{W}_1 * \mathbf{x}_1 - \mathbf{W}_2 * \mathbf{x}_2 - \mathbf{W}_3 * \mathbf{x}_3 - \mathbf{W}_4 * \mathbf{x}_4 - \mathbf{W}_5 * \mathbf{x}_5 - \mathbf{W}_6 * \mathbf{x}_6 - \mathbf{W}_7 * \mathbf{x}_7) \\
&\quad + (\mathbf{W}_0 * \mathbf{x}_1 + \mathbf{W}_1 * \mathbf{x}_0 + \mathbf{W}_2 * \mathbf{x}_3 - \mathbf{W}_3 * \mathbf{x}_2 + \mathbf{W}_4 * \mathbf{x}_5 - \mathbf{W}_5 * \mathbf{x}_4 - \mathbf{W}_6 * \mathbf{x}_7 + \mathbf{W}_7 * \mathbf{x}_6) e_1 \\
&\quad + (\mathbf{W}_0 * \mathbf{x}_2 - \mathbf{W}_1 * \mathbf{x}_3 + \mathbf{W}_2 * \mathbf{x}_0 + \mathbf{W}_3 * \mathbf{x}_1 + \mathbf{W}_4 * \mathbf{x}_6 + \mathbf{W}_5 * \mathbf{x}_7 - \mathbf{W}_6 * \mathbf{x}_4 - \mathbf{W}_7 * \mathbf{x}_5) e_2 \\
&\quad + (\mathbf{W}_0 * \mathbf{x}_3 + \mathbf{W}_1 * \mathbf{x}_2 - \mathbf{W}_2 * \mathbf{x}_1 + \mathbf{W}_3 * \mathbf{x}_0 + \mathbf{W}_4 * \mathbf{x}_7 - \mathbf{W}_5 * \mathbf{x}_6 + \mathbf{W}_6 * \mathbf{x}_5 - \mathbf{W}_7 * \mathbf{x}_4) e_3 \\
&\quad + (\mathbf{W}_0 * \mathbf{x}_4 - \mathbf{W}_1 * \mathbf{x}_5 - \mathbf{W}_2 * \mathbf{x}_6 - \mathbf{W}_3 * \mathbf{x}_7 + \mathbf{W}_4 * \mathbf{x}_0 + \mathbf{W}_5 * \mathbf{x}_1 + \mathbf{W}_6 * \mathbf{x}_2 + \mathbf{W}_7 * \mathbf{x}_3) e_4 \\
&\quad + (\mathbf{W}_0 * \mathbf{x}_5 + \mathbf{W}_1 * \mathbf{x}_4 + \mathbf{W}_3 * \mathbf{x}_6 - \mathbf{W}_2 * \mathbf{x}_7 - \mathbf{W}_4 * \mathbf{x}_1 + \mathbf{W}_5 * \mathbf{x}_0 - \mathbf{W}_6 * \mathbf{x}_3 + \mathbf{W}_7 * \mathbf{x}_2) e_5 \\
&\quad + (\mathbf{W}_0 * \mathbf{x}_6 + \mathbf{W}_1 * \mathbf{x}_7 + \mathbf{W}_2 * \mathbf{x}_4 - \mathbf{W}_3 * \mathbf{x}_5 - \mathbf{W}_4 * \mathbf{x}_2 + \mathbf{W}_5 * \mathbf{x}_3 + \mathbf{W}_6 * \mathbf{x}_0 - \mathbf{W}_7 * \mathbf{x}_1) e_6 \\
&\quad + (\mathbf{W}_0 * \mathbf{x}_7 - \mathbf{W}_1 * \mathbf{x}_6 + \mathbf{W}_2 * \mathbf{x}_5 + \mathbf{W}_3 * \mathbf{x}_4 - \mathbf{W}_4 * \mathbf{x}_3 - \mathbf{W}_5 * \mathbf{x}_2 + \mathbf{W}_6 * \mathbf{x}_1 + \mathbf{W}_7 * \mathbf{x}_0) e_7,
\end{aligned} \quad (5)$$

which can be expressed as the real matrix form as follows:

$$\begin{bmatrix} \mathcal{R}(\mathbf{W}_o *_o \mathbf{x}_o) \\ \mathcal{I}(\mathbf{W}_o *_o \mathbf{x}_o) \\ \mathcal{J}(\mathbf{W}_o *_o \mathbf{x}_o) \\ \mathcal{K}(\mathbf{W}_o *_o \mathbf{x}_o) \\ \mathcal{E}(\mathbf{W}_o *_o \mathbf{x}_o) \\ \mathcal{L}(\mathbf{W}_o *_o \mathbf{x}_o) \\ \mathcal{M}(\mathbf{W}_o *_o \mathbf{x}_o) \\ \mathcal{N}(\mathbf{W}_o *_o \mathbf{x}_o) \end{bmatrix} = \begin{bmatrix} \mathbf{W}_0 & -\mathbf{W}_1 & -\mathbf{W}_2 & -\mathbf{W}_3 & -\mathbf{W}_4 & -\mathbf{W}_5 & -\mathbf{W}_6 & -\mathbf{W}_7 \\ \mathbf{W}_1 & \mathbf{W}_0 & -\mathbf{W}_3 & \mathbf{W}_2 & -\mathbf{W}_5 & \mathbf{W}_4 & \mathbf{W}_7 & -\mathbf{W}_6 \\ \mathbf{W}_2 & \mathbf{W}_3 & \mathbf{W}_0 & -\mathbf{W}_1 & -\mathbf{W}_6 & -\mathbf{W}_7 & \mathbf{W}_4 & \mathbf{W}_5 \\ \mathbf{W}_3 & -\mathbf{W}_2 & \mathbf{W}_1 & \mathbf{W}_0 & -\mathbf{W}_7 & \mathbf{W}_6 & -\mathbf{W}_5 & \mathbf{W}_4 \\ \mathbf{W}_4 & \mathbf{W}_5 & \mathbf{W}_6 & \mathbf{W}_7 & \mathbf{W}_0 & -\mathbf{W}_1 & -\mathbf{W}_2 & -\mathbf{W}_3 \\ \mathbf{W}_5 & -\mathbf{W}_4 & \mathbf{W}_7 & -\mathbf{W}_6 & \mathbf{W}_1 & \mathbf{W}_0 & \mathbf{W}_3 & -\mathbf{W}_2 \\ \mathbf{W}_6 & -\mathbf{W}_7 & -\mathbf{W}_4 & \mathbf{W}_5 & \mathbf{W}_2 & -\mathbf{W}_3 & \mathbf{W}_0 & \mathbf{W}_1 \\ \mathbf{W}_7 & \mathbf{W}_6 & -\mathbf{W}_5 & -\mathbf{W}_4 & \mathbf{W}_3 & \mathbf{W}_2 & -\mathbf{W}_1 & \mathbf{W}_0 \end{bmatrix} * \begin{bmatrix} \mathbf{x}_0 \\ \mathbf{x}_1 \\ \mathbf{x}_2 \\ \mathbf{x}_3 \\ \mathbf{x}_4 \\ \mathbf{x}_5 \\ \mathbf{x}_6 \\ \mathbf{x}_7 \end{bmatrix}, \quad (6)$$

where $*_o$ and $*$ denote octonion convolution and real convolution, respectively. $\mathbf{x}_i \in R^N$ and $\mathbf{W}_i \in R^{N \times N}$ with $i=1,2,\ldots,7$. $\mathcal{R}(\bullet)$ denotes the real part of $\bullet$, $\mathcal{I}(\bullet)$, $\mathcal{J}(\bullet)$, $\mathcal{K}(\bullet)$, $\mathcal{E}(\bullet)$, $\mathcal{L}(\bullet)$, $\mathcal{M}(\bullet)$ and $\mathcal{N}(\bullet)$ denote the seven different imaginary parts of $\bullet$, respectively. The implementation of octonion convolutional operation is shown in Fig. 2, where $M_\mathbf{r}, M_\mathbf{i}, M_\mathbf{j}, M_\mathbf{k}, M_\mathbf{e}, M_\mathbf{l}, M_\mathbf{m}, M_\mathbf{n}$ refer to eight parts of feature maps and $K_\mathbf{r}, K_\mathbf{i}, K_\mathbf{j}, K_\mathbf{k}, K_\mathbf{e}, K_\mathbf{l}, K_\mathbf{m}, K_\mathbf{n}$ refer to eight parts of kernels, and $M_{\mathbf{p}_1} * K_{\mathbf{p}_2} (\mathbf{p}_1, \mathbf{p}_2 = \mathbf{r}, \mathbf{i}, \mathbf{j}, \mathbf{k}, \mathbf{e}, \mathbf{l}, \mathbf{m}, \mathbf{n})$ refer to the result of a real convolution between the feature maps and the kernels. The real representations of complex convolution, quaternion convolution, and octonion convolution are shown in Appendix 1.

*3.3 Octonion batch normalization module*

Batch normalization [31] can accelerate deep network training by reducing internal covariate shift. It allows us to use much higher learning rates and be less careful about initialization. When applying the batch normalization to real numbers, it is sufficient to translate and scale these numbers such that their mean is zero and their variance is one. However, when applying the batch normalization to complex or quaternion numbers, it can not ensure equal variance in both the real and imaginary components. In order to overcome this problem, a whitening approach is used in [65, 67], which scales the data by the square root of their variances along each principle components. In this section, we use the similar approach, but treating this problem as a "whitening" 8D vector problem.

Firstly, whitening is accomplished by multiplying the zero-centered data (**x**-E[**x**]) by the inverse square root of the covariance matrix **V**:

$$\tilde{\mathbf{x}} = \frac{(\mathbf{x} - \mathrm{E}[\mathbf{x}])}{\sqrt{\mathbf{V}}}, \quad (7)$$

and

$$\mathbf{V} = \begin{bmatrix} V_{rr} & V_{ri} & V_{rj} & V_{rk} & V_{re} & V_{rl} & V_{rm} & V_{rn} \\ V_{ir} & V_{ii} & V_{ij} & V_{ik} & V_{ie} & V_{il} & V_{im} & V_{in} \\ V_{jr} & V_{ji} & V_{jj} & V_{jk} & V_{je} & V_{jl} & V_{jm} & V_{jn} \\ V_{kr} & V_{ki} & V_{kj} & V_{kk} & V_{ke} & V_{kl} & V_{km} & V_{kn} \\ V_{er} & V_{ei} & V_{ej} & V_{ek} & V_{ee} & V_{el} & V_{em} & V_{en} \\ V_{lr} & V_{li} & V_{lj} & V_{lk} & V_{le} & V_{ll} & V_{lm} & V_{ln} \\ V_{mr} & V_{mi} & V_{mj} & V_{mk} & V_{me} & V_{ml} & V_{mm} & V_{mn} \\ V_{nr} & V_{ni} & V_{nj} & V_{nk} & V_{ne} & V_{nl} & V_{nm} & V_{nn} \end{bmatrix}, \quad (8)$$

where E[**x**] refers to the average value of each batch of training data $\mathbf{x} \in O^N$, and $\mathbf{V} \in O^{8 \times 8}$ is the covariance matrix of each batch of data **x**.

In order to avoid calculating the $\left(\sqrt{\mathbf{V}}\right)^{-1}$, Eq. (7) can be computed as follows

$$\tilde{\mathbf{x}} = \mathbf{U}(\mathbf{x} - \mathrm{E}[\mathbf{x}]), \quad (9)$$

where **U** is one of the matrices from the Cholesky decomposition of $\mathbf{V}^{-1}$, and each item of the matrix **U** is shown in Appendix 2.

Secondly, the forward conduction formula of the octonion batch normalization layer is defined as

$$\mathrm{OctonionBN}(\tilde{\mathbf{x}}) = \boldsymbol{\gamma}\tilde{\mathbf{x}} + \boldsymbol{\beta}, \quad (10)$$

where $\boldsymbol{\beta} = E(\mathbf{x}) \in O^8$ is a learned parameter with eight real parameters (one real part and seven imaginary parts) and $\boldsymbol{\gamma} = \sqrt{\mathbf{V}} \in O^{8 \times 8}$ is also a learned parameter with only 36 independent real parameters,

$$\boldsymbol{\gamma} = \begin{pmatrix} \gamma_{rr} & \gamma_{ri} & \gamma_{rj} & \gamma_{rk} & \gamma_{re} & \gamma_{rl} & \gamma_{rm} & \gamma_{rn} \\ \gamma_{ir} & \gamma_{ii} & \gamma_{ij} & \gamma_{ik} & \gamma_{ie} & \gamma_{il} & \gamma_{im} & \gamma_{in} \\ \gamma_{jr} & \gamma_{ji} & \gamma_{jj} & \gamma_{jk} & \gamma_{je} & \gamma_{jl} & \gamma_{jm} & \gamma_{jn} \\ \gamma_{kr} & \gamma_{ki} & \gamma_{kj} & \gamma_{kk} & \gamma_{ke} & \gamma_{kl} & \gamma_{km} & \gamma_{kn} \\ \gamma_{er} & \gamma_{ei} & \gamma_{ej} & \gamma_{ek} & \gamma_{ee} & \gamma_{el} & \gamma_{em} & \gamma_{en} \\ \gamma_{lr} & \gamma_{li} & \gamma_{lj} & \gamma_{lk} & \gamma_{le} & \gamma_{ll} & \gamma_{lm} & \gamma_{ln} \\ \gamma_{mr} & \gamma_{mi} & \gamma_{mj} & \gamma_{mk} & \gamma_{me} & \gamma_{ml} & \gamma_{mm} & \gamma_{mn} \\ \gamma_{nr} & \gamma_{ni} & \gamma_{nj} & \gamma_{nk} & \gamma_{ne} & \gamma_{nl} & \gamma_{nm} & \gamma_{nn} \end{pmatrix}. \quad (11)$$

Similar to [65] and [67], the diagonal of $\boldsymbol{\gamma}$ is initialized to $1/\sqrt{8}$, the off diagonal terms of $\boldsymbol{\gamma}$ and all components of $\boldsymbol{\beta}$ are initialized to 0.

### 3.4 Octonion weight initialization method

Before starting to train the network, we need to initialize the parameters of the network. If the weights are initialized to the same value, the updated weights are the same which means that the network can't learn the features. For deep neural networks, such initialization will make deeper meaningless and will not match the effects of linear classifiers. Therefore, the initial weight value is close but not equal to 0, that is, we initialize the weights to a small value, which not only ensures the difference between the input and output, but also allows the model to converge stably and quickly. In view of this, we provide an initialization method for octonion weight.

First, the octonion weight $\mathbf{W}_o \in O^{N \times N}$ is expressed as a polar form:

$$\mathbf{W}_o = |\mathbf{W}_o| e^{s\psi} = \mathcal{R}(\mathbf{W}_o) + \mathcal{I}(\mathbf{W}_o)e_1 + \mathcal{J}(\mathbf{W}_o)e_2 + \mathcal{K}(\mathbf{W}_o)e_3 \\ + \mathcal{E}(\mathbf{W}_o)e_4 + \mathcal{L}(\mathbf{W}_o)e_5 + \mathcal{M}(\mathbf{W}_o)e_6 + \mathcal{N}(\mathbf{W}_o)e_7, \quad (12)$$

where $|\mathbf{W}_o|$ is the magnitude, $\psi$ is the octonion phase, $s = s_1e_1 + s_2e_2 + s_3e_3 + s_4e_4 + s_5e_5 + s_6e_6 + s_7e_7$ is defined as purely imaginary, and $\sum_{p=1}^{7} s_p^2 = 1$, and $s_p$, $p=1,2,\ldots,7$, are expressed as:

$$\begin{aligned} s_1 &= \cos\theta_1; \quad s_2 = \sin\theta_1 \cos\theta_2; \quad s_3 = \sin\theta_1 \sin\theta_2 \cos\theta_3; \quad s_4 = \sin\theta_1 \sin\theta_2 \sin\theta_3 \cos\theta_4; \quad s_5 = \sin\theta_1 \sin\theta_2 \sin\theta_3 \sin\theta_4 \cos\theta_5; \\ s_6 &= \sin\theta_1 \sin\theta_2 \sin\theta_3 \sin\theta_4 \sin\theta_5 \cos\theta_6; \quad s_7 = \sin\theta_1 \sin\theta_2 \sin\theta_3 \sin\theta_4 \sin\theta_5 \sin\theta_6. \end{aligned} \quad (13)$$

where $0 < \theta_1, \theta_2, \theta_3, \theta_4, \theta_5 < \pi$ and $0 < \theta_6 < 2\pi$.

Then, we want to get the variance of $\mathbf{W}_o$ as follows

$$Var(\mathbf{W}_o) = E[|\mathbf{W}_o|^2] - (E[\mathbf{W}_o])^2, \quad (14)$$

where $|\mathbf{W}_o|$ follows an independent normal distribution of eight degrees of freedom (DOFs),

$$Var(\mathbf{W}_o) = \begin{cases} 2/(n_{in} + n_{out}) & \text{if Glorot's initialization [76] is used} \\ 2/n_{in} & \text{if He's initialization [77] is used} \end{cases}. \quad (15)$$

When $\mathbf{W}_o$ is symmetrically distributed around 0, that is $E[\mathbf{W}_o]=0$, then (14) reduces to

$$E[|\mathbf{W}_o|^2] = \int_0^\infty x^2 f(x) dx, \quad (16)$$

where $f(x)$ is the probability density function

$$f(x) = f_R(x;\sigma) + f_I(x;\sigma)e_1 + f_J(x;\sigma)e_2 + f_K(x;\sigma)e_3 + f_E(x;\sigma)e_4 + f_L(x;\sigma)e_5 + f_M(x;\sigma)e_6 + f_N(x;\sigma)e_7, \quad (17)$$

whose eight parts (one real part and seven imaginary parts) are all normally distributed, centered at zero, and independent.

Now we can see that the octonion weight initialization problem is now converted to initialize the octonion magnitude $|\mathbf{W}_o|$ and the octonion phase $\psi$. The phase initialization can be simply using the mean distribution between $-\pi$ and $\pi$. The magnitude initialization is shown in Theorem 1.

**Theorem 1**. The magnitude initialization can be implemented by using the Rayleigh distribution of $\sigma$, where

$$\sigma = \begin{cases} 1/\left(2\sqrt{n_{in}+n_{out}}\right), & \text{if Glorot's initialization [76] is used} \\ 2/\sqrt{n_{in}}, & \text{if He's initialization [77] is used} \end{cases}. \tag{18}$$

Proof.

The density function of eight parts of $f(x)$ in (17) can be expressed as follows:

$$f_R(x;\sigma) = f_I(x;\sigma) = f_J(x;\sigma) = f_K(x;\sigma) = f_E(x;\sigma) = f_L(x;\sigma) = f_M(x;\sigma) = f_N(x;\sigma) = \frac{e^{-x^2/(2\sigma^2)}}{\sqrt{2\pi\sigma^2}}. \tag{19}$$

The length of $f(x)$ can be expressed as

$$q = \sqrt{f_R^2(x;\sigma) + f_I^2(x;\sigma) + f_J^2(x;\sigma) + f_K^2(x;\sigma) + f_E^2(x;\sigma) + f_L^2(x;\sigma) + f_M^2(x;\sigma) + f_N^2(x;\sigma)}, \tag{20}$$

and then the cumulative distribution function of length $q$ can be expressed as:

$$F_q(x;\sigma) = \iiint\iiint\iint_{H_x} f_R(x;\sigma)f_I(x;\sigma)f_J(x;\sigma)f_K(x;\sigma)f_E(x;\sigma)f_L(x;\sigma)f_M(x;\sigma)f_N(x;\sigma)\mathrm{d}H_x, \tag{21}$$

where $H_x$ is a hypersphere

$$H_x = \{(r,i,j,k,e,l,m,n) : \sqrt{r^2+i^2+j^2+k^2+e^2+l^2+m^2+n^2} < x\}, \tag{22}$$

and using (12) and (13), we have

$$\begin{aligned}\mathrm{d}H_x &= \mathrm{d}r\,\mathrm{d}i\,\mathrm{d}j\,\mathrm{d}k\,\mathrm{d}e\,\mathrm{d}l\,\mathrm{d}m\,\mathrm{d}n \\ &= \rho^7 \sin^6\phi_1 \sin^5\phi_2 \sin^4\phi_3 \sin^3\phi_4 \sin^2\phi_5 \sin\phi_6\,\mathrm{d}\rho\,\mathrm{d}\phi_1\,\mathrm{d}\phi_2\,\mathrm{d}\phi_3\,\mathrm{d}\phi_4\,\mathrm{d}\phi_5\,\mathrm{d}\phi_6\,\mathrm{d}\phi_7\end{aligned}, \tag{23}$$

where $\rho = \sqrt{r^2+i^2+j^2+k^2+e^2+l^2+m^2+n^2}$ indicates the size of the octonion number.

The calculation of (23) is shown in Appendix 3.

From (21), we have

$$F_q(x;\sigma) = \left(\frac{1}{\sqrt{2\pi\sigma^2}}\right)^8 \int\int\int\int\int\int\int\int_0^x e^{-r^2/2\sigma^2} e^{-i^2/2\sigma^2} e^{-j^2/2\sigma^2} e^{-k^2/2\sigma^2} e^{-e^2/2\sigma^2} e^{-l^2/2\sigma^2} e^{-m^2/2\sigma^2} e^{-n^2/2\sigma^2} \mathrm{d}H_x$$

$$= \frac{1}{16\pi^4\sigma^8} \int_0^{2\pi}\int_0^{\pi}\int_0^{\pi}\int_0^{\pi}\int_0^{\pi}\int_0^{\pi}\int_0^{\pi}\int_0^x e^{-\rho^2/2\sigma^2} \rho^7 \sin^6\phi_1 \sin^5\phi_2 \sin^4\phi_3 \sin^3\phi_4 \sin^2\phi_5 \sin\phi_6 \, \mathrm{d}\rho \, \mathrm{d}\phi_1 \, \mathrm{d}\phi_2 \, \mathrm{d}\phi_3 \, \mathrm{d}\phi_4 \, \mathrm{d}\phi_5 \, \mathrm{d}\phi_6 \, \mathrm{d}\phi_7$$

$$= \frac{1}{16\pi^4\sigma^8} \int_0^{2\pi} \mathrm{d}\phi_7 \int_0^{\pi} \sin\phi_6 \mathrm{d}\phi_6 \int_0^{\pi} \sin^2\phi_5 \mathrm{d}\phi_5 \int_0^{\pi} \sin^3\phi_4 \mathrm{d}\phi_4 \int_0^{\pi} \sin^4\phi_3 \mathrm{d}\phi_3 \int_0^{\pi} \sin^5\phi_2 \mathrm{d}\phi_2 \int_0^{\pi} \sin^6\phi_1 \mathrm{d}\phi_1 \int_0^x \rho^7 e^{-\rho^2/2\sigma^2} \mathrm{d}\rho$$

$$= \frac{1}{16\pi^4\sigma^8} \cdot 2\pi \cdot 2 \left[\frac{1}{2}\phi_5 - \frac{1}{4}\sin 2\phi_5\right]_0^{\pi} \left[\frac{1}{12}\cos 3\phi_4 - \frac{3}{4}\cos\phi_4\right]_0^{\pi} \left[\frac{1}{8}\sin 4\phi_3 + \frac{1}{16}\sin 2\phi_3 + \frac{3}{8}\phi_3\right]_0^{\pi}$$

$$\cdot \left[-\frac{1}{80}\cos 5\phi_2 + \frac{5}{48}\cos 3\phi_2 - \frac{5}{8}\cos\phi_2\right]_0^{\pi} \left[-\frac{1}{192}\sin 6\phi_1 + \frac{1}{8}\sin 4\phi_1 + \frac{15}{192}\sin 2\phi_1 + \frac{5}{16}\phi_1\right]_0^{\pi} \int_0^x \rho^7 e^{-\rho^2/2\sigma^2} \mathrm{d}\rho$$

$$= \frac{1}{16\pi^4\sigma^8} \cdot 2\pi \cdot 2 \cdot \frac{1}{2}\pi \cdot \frac{4}{3} \cdot \frac{3}{8}\pi \cdot \frac{16}{15} \cdot \frac{5}{16}\pi \cdot \int_0^x \rho^7 e^{-\rho^2/2\sigma^2} \mathrm{d}\rho$$

$$= \frac{1}{48\sigma^8} \int_0^x \rho^7 e^{-\rho^2/2\sigma^2} \mathrm{d}\rho.$$

(24)

Combining (16) and (24), and considering the probability density function $f(x)$ is the derivative of the cumulative distribution function $F_q(x;\sigma)$, we have

$$\mathrm{E}[|\mathbf{W}_o|^2] = \int_0^{\infty} x^2 f(x) \mathrm{d}x = \int_0^{\infty} x^2 \mathrm{d}F_q(x;\sigma) = \frac{1}{48\sigma^8} \int_0^{\infty} x^9 e^{-x^2/2\sigma^2} \mathrm{d}x.$$ (25)

The proof of (25) is shown in Appendix 4.

With integration by parts and the L'Hôpital's rule, we obtain

$$\mathrm{E}[|\mathbf{W}_o|^2] = 8\left(\int_0^{\infty} x e^{-x^2/2\sigma^2} \mathrm{d}x\right) = 8\sigma^2.$$ (26)

Now we have the variance of $W$ expressed in terms of a single parameter $\sigma$:

$$Var(\mathbf{W}_o) = 8\sigma^2 = \begin{cases} 2/(n_{in} + n_{out}) & \text{if Glorot's initialization [76] is used} \\ 2/n_{in} & \text{if He's initialization [77] is used} \end{cases}$$ (27)

From (27), we can obtain (14).  □

## 4. Experimental Results and Discussion

Similar to the 110-layer deep real networks [25], we design an octonion convolutional neural network named deep octonion networks, whose schematic diagram are shown in Fig. 3. Then we performed the image classification tasks of CIFAR-10 and CIFAR-100 [66] to verify the validity of the proposed deep octonion networks. The following experiment is implemented using Keras (Tensorflow as backend) on a PC machine, which sets up Ubuntu 16.04 operating system and has an Intel(R) Core(TM) i7-2600 CPU with speed of 3.40 GHz and 64 GB RAM, and has also two NVIDIA GeForce GTX1080-Ti GPUs.

4.1 Models configurations

4.1.1 Octonion input construction

Since the images in datasets of CIFAR-10 and CIFAR-100 are real-valued, however, the input of the proposed deep octonion networks needs to be octonion matrix, which we should to be obtained first. The octonion has one real part and seven imaginary parts, we put the original $N$ training real images into the real part, and similar to [65] and [67], the seven imaginary parts of octonion matrix are obtained by performing a single real-valued residual block ( BN $\rightarrow$ ReLU $\rightarrow$ Conv $\rightarrow$ BN $\rightarrow$ ReLU $\rightarrow$ Conv ) [25] 7 times at the same time. Then, the 8 vectors are connected according to a given axis to form a brand new octonion vector.

4.1.2 The structure of deep octonion networks

The OctonionConv $\rightarrow$ OctonionBN $\rightarrow$ ReLU operation is performed on the obtained octonion input, where OctonionConv denotes the octonion convolution module shown in section 3.2 and OctonionBN denotes the octonion batch normalization module shown in section 3.3. Then the octonion output is sent to the next three stages. In each stage, there are several residual blocks with double convolution layers. The characteristics of the three stages are the same, and the number is progressively layer by layer to ensure the expressive ability of the output features. To speed up the training, the following layer is an AveragePooling2D layer, which is then followed by a fully connected layer called Dense to classify the input. The deep octonion network model sets the number of residual blocks in the three stages to 10, 9, and 9, respectively, and the number of convolution filters is set to 32, 64, and 128. The batch size is set to 64.

4.1.3 The training of deep octonion networks

Deep octonion networks are then compiled, the cross entropy loss function and the stochastic gradient descent method are chosen for training the model. Set the Nesterov Momentum to 0.9 in the back propagation of stochastic gradient descent to increase the stability and speed up the best advantage of reaching the global. The learning rate is shown in Table 2. Using a custom learning rate schedule, different learning rates are used in different epochs. Here, the learning rate from 0 to 80 epoch is divided into three stages. The first 20 epochs are preheated at a learning rate of 0.01, and the middle 40 epochs increase the learning rate by ten times. The latter 20 epoch will restore to 0.01. The learning rate from 80 epochs is set to make the network more stable. The deep octonion networks are trained 120 epochs, which are less than 200 epochs in [65] and [67], because the convergence speed of the deep octonion networks is higher than the deep real networks [25], deep complex networks [65], and deep quaternion networks [67].

4.2 Experimental results and analysis

There are two methods for choosing the learning rate. One is to have the same learning rate during training which is called "smooth" learning rate (blue line in Fig. 4), and the other is to adjust the learning rate during training which is called "convex" learning rate (red line in Fig. 4). Experiments show that the "convex" learning rate setting from epoch 0 to epoch 80 is better than the "smooth" learning rate. Although the accuracy will fluctuate during the 20 to 60 epochs, the accuracy will increase slightly in the next 20 epochs. Therefore, in the subsequent experiments, the "convex" learning rate was used for scheduling. The accuracy of the deep octonion networks, deep complex networks [65], and deep quaternion networks [67] in the first 20 epochs are compared in Fig. 5, from which we can see that the proposed deep octonion networks are better than other compared deep networks. Besides, the deep octonion networks have less learning parameters than other compared deep networks as shown in Table 3.

For the classification tasks of CIFAR-10 and CIFAR-100: Firstly, we use the 10-fold cross-validation method, that is, we divide the data set into ten parts, and take nine parts as training data and one part as testing data, and then the Top-1 error rate is obtained by averaging the 10 results. Secondly, the 10-fold cross-validation are performed 10 times, and the mean value is obtained as an estimation of the Top-1 error rate of the algorithm. Table 3 shows the Top-1 error rate of the deep octonion networks, deep real networks [25], deep complex networks [65], and deep quaternion networks [67] on CIFAR-10 and CIFAR-100. It can be seen from the Table 3 that the deep octonion networks can achieve lower error rate than other compared deep networks and the advantage becomes more apparent when there are more classes to be classified. From the second column of the Table 3, we can see that compared with deep quaternion networks [67], the improvement of the deep octonion networks is not significant, improved relatively 1.7%. However, as shown in the third column of Table 3, when the number of classes increases, the proposed deep octonion networks improve relatively 5.4% when compared to deep quaternion networks [67]. This also implies the deep octonion networks have better generalization ability than other compared deep networks. The phenomenon is also explained by the multi-tasking learning in the next section.

## 5. Explanation of deep octonion networks via multi-task learning

In Machine Learning, the standard algorithm is to learn one task at a time, we generally train a single model or an ensemble of models to perform our desired task and the output of the system is real. When face the complex learning problems, traditional methods chose a similar approach. Firstly, decompose the complex learning problems into simple and independent sub-problems, and then study each sub-problem separately. Finally, establish the mathematical model of complex problems by combining the sub-problem learning results. These operations seem reasonable but not accurate, for the reason that many problems in the real world cannot be decomposed into an independent sub-problem, the rich interrelated information contained between the problems cannot be ignored. Even if it can be decomposed, the sub-problems are interrelated, and connected by sharing factors or share representation. In order to solve this problem, multi-tasking learning (MTL) was born [73]. Compared with single-task learning (STL—learning just one task at a time), MTL is a kind of joint learning method which learns multiple related tasks together based on shared representations. The purpose of the shared representation is to improve the generalization. Multiple tasks are learned in parallel, and the results affect each other.

Figures 6(a) and 6(b) show a single-task learning and multi-task feedforward neural network with one input layer, two hidden layers and one output layer, respectively. Figure 6(c) shows that the DONs proposed in this paper which use a network structure similar to Figure 6(b) to learn eight related tasks together. The input of the last seven tasks is learned by the input of the first one. So there are related parts and irrelevant parts among the eight tasks. The details of the DONs shared layers follow the rules in Fig. 2.

It can be seen from Fig. 4 that the learning between the tasks is independent of each other in the single-task learning (Fig. 6(a)). In multi-task learning, parameters between multiple tasks are shared (Figure 6(b)). In the context of deep learning, multi-task learning is typically done with either hard or soft parameter sharing of hidden layers [74]. Figure 6 (b) and 6(c) present the multi-task learning mode of hard parameter sharing, we refer to the structure between the input layer and the output layer as the shared layer. Hard parameter sharing is the most commonly used approach to MTL in neural networks and greatly reduces the risk of overfitting [74]. It has been demonstrated in [75] that the risk of overfitting the shared parameters is smaller than overfitting the specific parameters for each task. What's more, regard the number of tasks as $N$, a larger $N$ means that the more tasks are learned simultaneously, the more the model can find a representation that captures all tasks, and the less likely the original task is to overfitting. This also explains that the performance can be improved when the relationships between convolution kernels are modeled by complex algebra, quaternion algebra, and also octonion algebra and why deep neural networks can achieve better results on octonion domains.

Thinking of DONs as a specific MTL. The neural network can help the hidden layer to escape local minimum values through the interaction between different tasks while learning. When learning a Main task, the part that is not relevant to the task acts as noise during the learning process. As different tasks have different noise patterns, a model that learns multi tasks simultaneously is able to learn more general representations [74]. Regard the first of the eight tasks of DONs as the main task, and the last seven as auxiliary tasks, it can improve the learning rate of shared representation, reduce network over-fitting, and improve generalization effect by multi-tasks parallel learning. The more similar two tasks benefit more in multi-task learning.

# 6. Conclusion

In this paper, we proposed deep octonion networks (DONs) as an extension of DRNs, the DCNs, and the DQNs. The main building blocks of DONs are also given, such as octonion convolution, octonion batch normalization, and octonion weight initialization. DONs are applied to the image classification tasks of CIFAR-10 and CIFAR-100 to verify the validity of the octonion network. Experiments showed that compared with the DRNs, the DCNs, and the DQNs, the proposed DONs have better convergence, less parameters, and higher classification accuracy. The success of DONs is also explained through multi-task learning.

**ACKNOWLEDGEMENT**

This work was supported in part by the National Natural Science Foundation of China under Grants 61876037, 31800825, 61871117, 61871124, 61773117, 31571001, 61572258, and in part by the National Key Research and Development Program of China under Grant 2017YFC0107900, and in part by the Short-Term Recruitment Program of Foreign Experts under Grant WQ20163200398.

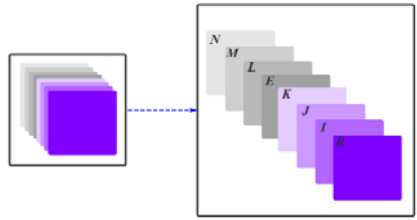

**Fig. 1.** The octonion internal representation

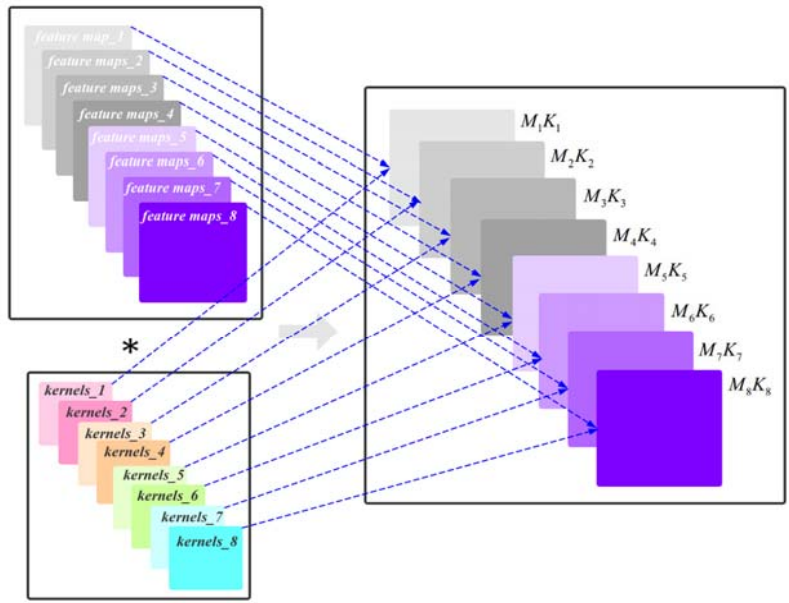

(a) Real Convolution

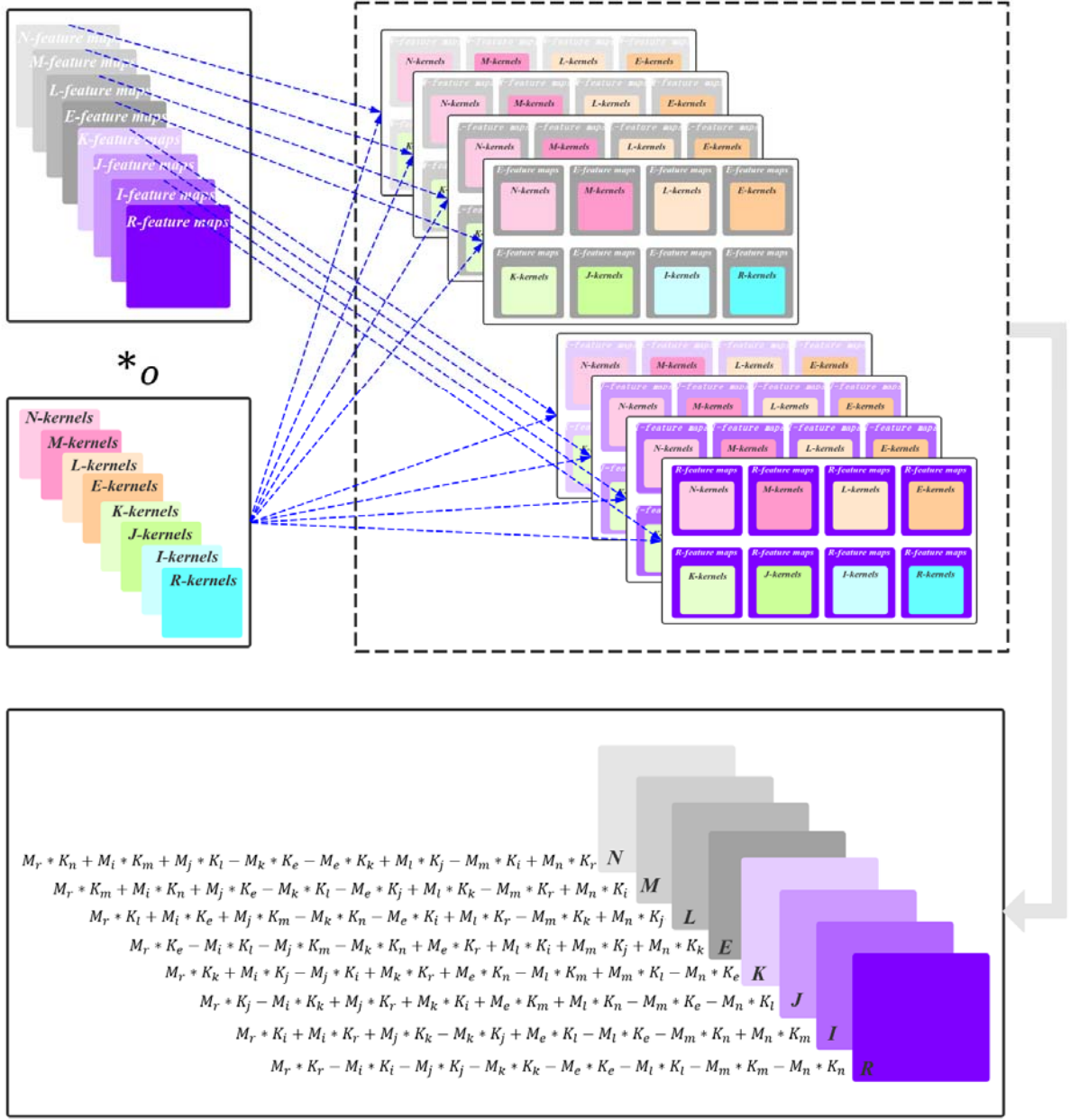

**Fig. 2.** Illustration of the real convolution (a) and octonion convolution (b)

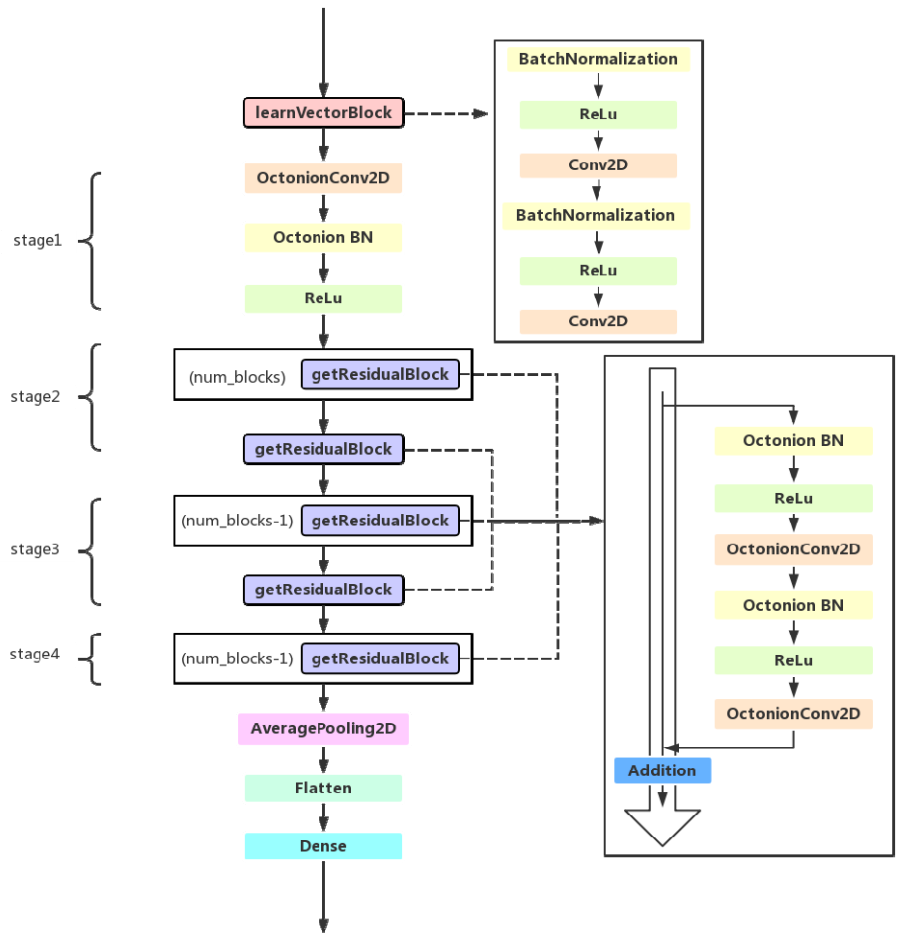

**Fig. 3.** Structure diagram of deep octonion networks

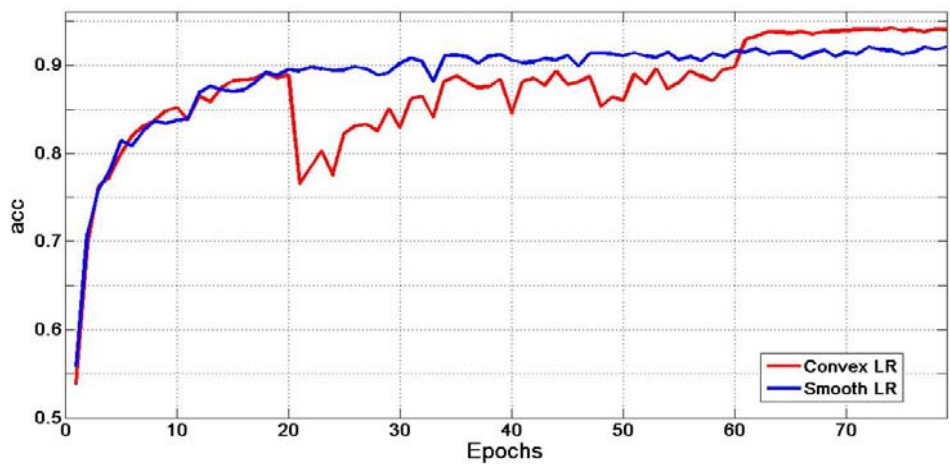

**Fig. 4.** Results of different learning rate settings

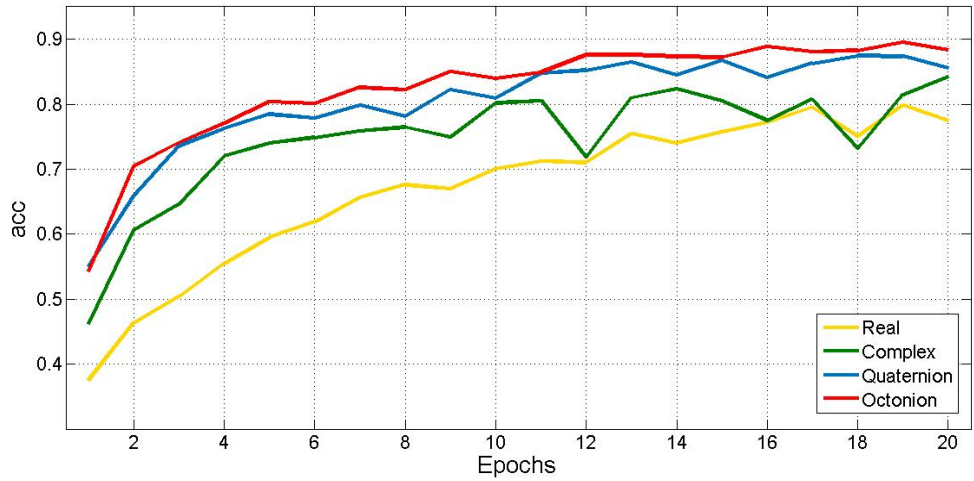

**Fig. 5.** Accuracy curves of four models in the first 20 epochs

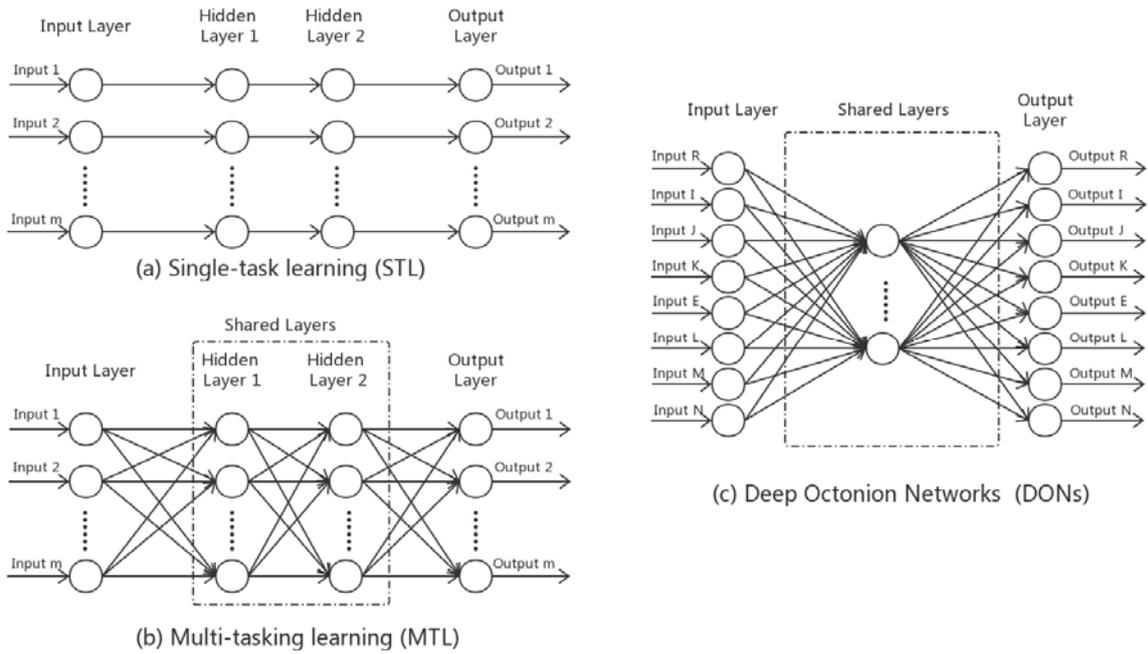

**Fig. 6.** The comparison of STL, MTL and DONs

**Table 1**
The multiplication table of the unit octonions

| | $e_i e_j$ | $e_j$ | | | | | | | |
|---|---|---|---|---|---|---|---|---|---|
| | | 1 | $e_1$ | $e_2$ | $e_3$ | $e_4$ | $e_5$ | $e_6$ | $e_7$ |
| $e_i$ | 1 | 1 | $e_1$ | $e_2$ | $e_3$ | $e_4$ | $e_5$ | $e_6$ | $e_7$ |
| | $e_1$ | $e_1$ | $-1$ | $e_3$ | $-e_2$ | $e_5$ | $-e_4$ | $-e_7$ | $e_6$ |
| | $e_2$ | $e_2$ | $-e_3$ | $-1$ | $e_1$ | $e_6$ | $e_7$ | $-e_4$ | $-e_5$ |
| | $e_3$ | $e_3$ | $e_2$ | $-e_1$ | $-1$ | $e_7$ | $-e_6$ | $e_5$ | $-e_4$ |
| | $e_4$ | $e_4$ | $-e_5$ | $-e_6$ | $-e_7$ | $-1$ | $e_1$ | $e_2$ | $e_3$ |
| | $e_5$ | $e_5$ | $e_4$ | $-e_7$ | $e_6$ | $-e_1$ | $-1$ | $-e_3$ | $e_2$ |
| | $e_6$ | $e_6$ | $e_7$ | $e_4$ | $-e_5$ | $-e_2$ | $e_3$ | $-1$ | $-e_1$ |
| | $e_7$ | $e_7$ | $-e_6$ | $e_5$ | $e_4$ | $-e_3$ | $-e_2$ | $e_1$ | $-1$ |

**Table 2**
The learning rate (%) of octonion convolution neural network.

| Epoch | Learning-rate |
|---|---|
| (0 , 20) | 0.01 |
| (20, 60) | 0.1 |
| (60, 80) | 0.01 |
| (80,110) | 0.001 |
| (110, 120) | 0.0001 |

**Table 3**
The classification error rate of three models in two types of datasets

| Architecture | Params | CIFAR-10 | CIFAR-100 |
|---|---|---|---|
| Real [25] | 3,619,844 | 6.37 | - |
| Complex [65] | 1,823,620 | 5.60 | 27.09 |
| Quaternion [67] | 932,792 | 5.44 | 26.01 |
| Octonion | **481,150** | **5.35** | **24.60** |

**Appendix 1** The real representations of complex convolution, quaternion convolution, and octonion convolution

| Multidimensional domains | Input vector | Filter matrix | Various convolutions between **W** and **h** | Implementation of various convolutions in real domain |
|---|---|---|---|---|
| Real [25] | $\mathbf{x}_r$ | $\mathbf{W}_r$ | $\mathbf{W}_r * \mathbf{x}_r = \sum_i \sum_j W_r[i,j] x_r[m-i, n-j]$ | $\mathcal{R}(\mathbf{W} * \mathbf{x}) = \mathbf{W} * \mathbf{x}$ |
| Complex [65] | $\mathbf{x}_c = \mathbf{x}_0 + \mathbf{x}_1 e_1$ | $\mathbf{W}_c = \mathbf{W}_0 + \mathbf{W}_1 e_1$ | $\mathbf{W}_c *_c \mathbf{x}_c = (\mathbf{W}_0 + \mathbf{W}_1 e_1) *_c (\mathbf{x}_0 + \mathbf{x}_1 e_1)$ | $\begin{bmatrix} \mathcal{R}(\mathbf{W}_c *_c \mathbf{x}_c) \\ \mathcal{I}(\mathbf{W}_c *_c \mathbf{x}_c) \end{bmatrix} = \begin{bmatrix} \mathbf{W}_0 & -\mathbf{W}_1 \\ \mathbf{W}_1 & \mathbf{W}_0 \end{bmatrix} * \begin{bmatrix} \mathbf{x}_0 \\ \mathbf{x}_1 \end{bmatrix}$ |
| Quaternion [67] | $\mathbf{x}_q = \mathbf{x}_0 + \mathbf{x}_1 e_1 + \mathbf{x}_2 e_2 + \mathbf{x}_3 e_3$ | $\mathbf{W}_q = \mathbf{W}_0 + \mathbf{W}_1 e_1 + \mathbf{W}_2 e_2 + \mathbf{W}_3 e_3$ | $\mathbf{W}_q *_q \mathbf{x}_q = (\mathbf{W}_0 + \mathbf{W}_1 e_1 + \mathbf{W}_2 e_2 + \mathbf{W}_3 e_3) *_q (\mathbf{x}_0 + \mathbf{x}_1 e_1 + \mathbf{x}_2 e_2 + \mathbf{x}_3 e_3)$ | $\begin{bmatrix} \mathcal{R}(\mathbf{W}_q *_q \mathbf{x}_q) \\ \mathcal{I}(\mathbf{W}_q *_q \mathbf{x}_q) \\ \mathcal{J}(\mathbf{W}_q *_q \mathbf{x}_q) \\ \mathcal{K}(\mathbf{W}_q *_q \mathbf{x}_q) \end{bmatrix} = \begin{bmatrix} \mathbf{W}_0 & -\mathbf{W}_1 & -\mathbf{W}_2 & -\mathbf{W}_3 \\ \mathbf{W}_1 & \mathbf{W}_0 & -\mathbf{W}_3 & \mathbf{W}_2 \\ \mathbf{W}_2 & \mathbf{W}_3 & \mathbf{W}_0 & -\mathbf{W}_1 \\ \mathbf{W}_3 & -\mathbf{W}_2 & \mathbf{W}_1 & \mathbf{W}_0 \end{bmatrix} * \begin{bmatrix} \mathbf{x}_0 \\ \mathbf{x}_1 \\ \mathbf{x}_2 \\ \mathbf{x}_3 \end{bmatrix}$ |
| Octonion | $\mathbf{x}_o = \mathbf{x}_0 + \mathbf{x}_1 e_1 + \mathbf{x}_2 e_2 + \mathbf{x}_3 e_3 + \mathbf{x}_4 e_4 + \mathbf{x}_5 e_5 + \mathbf{x}_6 e_6 + \mathbf{x}_7 e_7$ | $\mathbf{W}_o = \mathbf{W}_0 + \mathbf{W}_1 e_1 + \mathbf{W}_2 e_2 + \mathbf{W}_3 e_3 + \mathbf{W}_4 e_4 + \mathbf{W}_5 e_5 + \mathbf{W}_6 e_6 + \mathbf{W}_7 e_7$ | $\mathbf{W}_o *_o \mathbf{x}_o = (\mathbf{W}_0 + \mathbf{W}_1 e_1 + \mathbf{W}_2 e_2 + \mathbf{W}_3 e_3 + \mathbf{W}_4 e_4 + \mathbf{W}_5 e_5 + \mathbf{W}_6 e_6 + \mathbf{W}_7 e_7) *_o (\mathbf{x}_0 + \mathbf{x}_1 e_1 + \mathbf{x}_2 e_2 + \mathbf{x}_3 e_3 + \mathbf{x}_4 e_4 + \mathbf{x}_5 e_5 + \mathbf{x}_6 e_6 + \mathbf{x}_7 e_7)$ | $\begin{bmatrix} \mathcal{R}(\mathbf{W}_o *_o \mathbf{x}_o) \\ \mathcal{I}(\mathbf{W}_o *_o \mathbf{x}_o) \\ \mathcal{J}(\mathbf{W}_o *_o \mathbf{x}_o) \\ \mathcal{K}(\mathbf{W}_o *_o \mathbf{x}_o) \\ \mathcal{E}(\mathbf{W}_o *_o \mathbf{x}_o) \\ \mathcal{L}(\mathbf{W}_o *_o \mathbf{x}_o) \\ \mathcal{M}(\mathbf{W}_o *_o \mathbf{x}_o) \\ \mathcal{N}(\mathbf{W}_o *_o \mathbf{x}_o) \end{bmatrix} = \begin{bmatrix} \mathbf{W}_0 & -\mathbf{W}_1 & -\mathbf{W}_2 & -\mathbf{W}_3 & -\mathbf{W}_4 & -\mathbf{W}_5 & -\mathbf{W}_6 & -\mathbf{W}_7 \\ \mathbf{W}_1 & \mathbf{W}_0 & -\mathbf{W}_3 & \mathbf{W}_2 & -\mathbf{W}_5 & \mathbf{W}_4 & \mathbf{W}_7 & -\mathbf{W}_6 \\ \mathbf{W}_2 & \mathbf{W}_3 & \mathbf{W}_0 & -\mathbf{W}_1 & -\mathbf{W}_6 & -\mathbf{W}_7 & \mathbf{W}_4 & \mathbf{W}_5 \\ \mathbf{W}_3 & -\mathbf{W}_2 & \mathbf{W}_1 & \mathbf{W}_0 & -\mathbf{W}_7 & \mathbf{W}_6 & -\mathbf{W}_5 & \mathbf{W}_4 \\ \mathbf{W}_4 & \mathbf{W}_5 & \mathbf{W}_6 & \mathbf{W}_7 & \mathbf{W}_0 & -\mathbf{W}_1 & -\mathbf{W}_2 & -\mathbf{W}_3 \\ \mathbf{W}_5 & -\mathbf{W}_4 & \mathbf{W}_7 & -\mathbf{W}_6 & \mathbf{W}_1 & \mathbf{W}_0 & \mathbf{W}_3 & -\mathbf{W}_2 \\ \mathbf{W}_6 & -\mathbf{W}_7 & -\mathbf{W}_4 & \mathbf{W}_5 & \mathbf{W}_2 & -\mathbf{W}_3 & \mathbf{W}_0 & \mathbf{W}_1 \\ \mathbf{W}_7 & \mathbf{W}_6 & -\mathbf{W}_5 & -\mathbf{W}_4 & \mathbf{W}_3 & \mathbf{W}_2 & -\mathbf{W}_1 & \mathbf{W}_0 \end{bmatrix} * \begin{bmatrix} \mathbf{x}_0 \\ \mathbf{x}_1 \\ \mathbf{x}_2 \\ \mathbf{x}_3 \\ \mathbf{x}_4 \\ \mathbf{x}_5 \\ \mathbf{x}_6 \\ \mathbf{x}_7 \end{bmatrix}$ |

$e$ denotes imaginary component, where $e_i^2 = -1$, $e_i e_j = -e_j e_i \neq e_j e_i$, $(e_i e_j) e_k = -e_i (e_j e_k) \neq e_i e_j e_k$, $\forall i \neq j \neq k$, $1 \leq i, j, k \leq 7$. $*$ denotes ordinary real convolution, $*_c$ denotes complex convolution, $*_q$ denotes quaternion convolution, $*_o$ denotes octonion convolution. $\mathcal{R}(\bullet)$ denotes the real components of $\bullet$, $\mathcal{I}(\bullet)$, $\mathcal{J}(\bullet)$, $\mathcal{K}(\bullet)$, $\mathcal{E}(\bullet)$, $\mathcal{L}(\bullet)$, $\mathcal{M}(\bullet)$ and $\mathcal{N}(\bullet)$ denote the different imaginary components of $\bullet$ respectively. Where $\mathbf{x}_r \in R^N$, $\mathbf{x}_c \in C^N$, $\mathbf{x}_q \in Q^N$, $\mathbf{x}_o \in O^N$, $\mathbf{W}_r \in R^{N \times N}$, $\mathbf{W}_c \in C^{N \times N}$, $\mathbf{W}_q \in Q^N$, $\mathbf{W}_o \in O^N$, $\mathbf{x}_i \in R^N, \mathbf{W}_i \in R^{N \times N}, i = 0, 1, ..., 7$, where $R$, $C$, $Q$, $O$ denotes real, complex, quaternion, octonion domain, respectively.

**Appendix 2** The matrix **U**:

$$\mathbf{U}_{rr} = \sqrt{V_{rr}}; \quad \mathbf{U}_{ri} = \frac{1}{\mathbf{U}_{rr}} * V_{ri}; \quad \mathbf{U}_{rj} = \frac{1}{\mathbf{U}_{rr}} * V_{rj}; \quad \mathbf{U}_{rk} = \frac{1}{\mathbf{U}_{rr}} * V_{rk}; \quad \mathbf{U}_{re} = \frac{1}{\mathbf{U}_{rr}} * V_{re}; \quad \mathbf{U}_{rl} = \frac{1}{\mathbf{U}_{rr}} * V_{rl}; \quad \mathbf{U}_{rm} = \frac{1}{\mathbf{U}_{rr}} * V_{rm}; \quad \mathbf{U}_{rn} = \frac{1}{\mathbf{U}_{rr}} * V_{rn};$$

$$\mathbf{U}_{ii} = \sqrt{V_{ii} - \mathbf{U}_{ri}^2}; \quad \mathbf{U}_{ij} = \frac{1}{\mathbf{U}_{ii}} * (V_{ij} - (\mathbf{U}_{ri} * \mathbf{U}_{rj})); \quad \mathbf{U}_{ik} = \frac{1}{\mathbf{U}_{ii}} * (V_{ik} - (\mathbf{U}_{rj} * \mathbf{U}_{rk})); \quad \mathbf{U}_{ie} = \frac{1}{\mathbf{U}_{ii}} * (V_{ie} - (\mathbf{U}_{rk} * \mathbf{U}_{re})); \quad \mathbf{U}_{il} = \frac{1}{\mathbf{U}_{ii}} * (V_{il} - (\mathbf{U}_{re} * \mathbf{U}_{rl}));$$

$$\mathbf{U}_{im} = \frac{1}{\mathbf{U}_{ii}} * (V_{im} - (\mathbf{U}_{rl} * \mathbf{U}_{rm})); \quad \mathbf{U}_{in} = \frac{1}{\mathbf{U}_{ii}} * (V_{in} - (\mathbf{U}_{rm} * \mathbf{U}_{rn}))$$

$$\mathbf{U}_{jj} = \sqrt{V_{jj} - (\mathbf{U}_{ij}^2 + \mathbf{U}_{rj}^2)}; \quad \mathbf{U}_{jk} = \frac{1}{\mathbf{U}_{jj}} * (V_{jk} - (\mathbf{U}_{ij} * \mathbf{U}_{ik} + \mathbf{U}_{rj} * \mathbf{U}_{rk})); \quad \mathbf{U}_{je} = \frac{1}{\mathbf{U}_{jj}} * (V_{je} - (\mathbf{U}_{ik} * \mathbf{U}_{ie} + \mathbf{U}_{rk} * \mathbf{U}_{re})); \quad \mathbf{U}_{jl} = \frac{1}{\mathbf{U}_{jj}} * (V_{jl} - (\mathbf{U}_{ie} * \mathbf{U}_{il} + \mathbf{U}_{re} * \mathbf{U}_{rl}));$$

$$\mathbf{U}_{jm} = \frac{1}{\mathbf{U}_{jj}} * (V_{jm} - (\mathbf{U}_{il} * \mathbf{U}_{im} + \mathbf{U}_{rl} * \mathbf{U}_{rm})); \quad \mathbf{U}_{jn} = \frac{1}{\mathbf{U}_{jj}} * (V_{jn} - (\mathbf{U}_{im} * \mathbf{U}_{in} + \mathbf{U}_{rm} * \mathbf{U}_{rn}));$$

$$\mathbf{U}_{kk} = \sqrt{V_{kk} - (\mathbf{U}_{jk}^2 + \mathbf{U}_{ik}^2 + \mathbf{U}_{rk}^2)}; \quad \mathbf{U}_{ke} = \frac{1}{\mathbf{U}_{kk}} * (V_{ke} - (\mathbf{U}_{jk} * \mathbf{U}_{je} + \mathbf{U}_{ik} * \mathbf{U}_{ie} + \mathbf{U}_{rk} * \mathbf{U}_{re})); \quad \mathbf{U}_{kl} = \frac{1}{\mathbf{U}_{kk}} * (V_{kl} - (\mathbf{U}_{je} * \mathbf{U}_{jl} + \mathbf{U}_{ie} * \mathbf{U}_{il} + \mathbf{U}_{re} * \mathbf{U}_{rl}));$$

$$\mathbf{U}_{km} = \frac{1}{\mathbf{U}_{kk}} * (V_{km} - (\mathbf{U}_{jl} * \mathbf{U}_{jm} + \mathbf{U}_{il} * \mathbf{U}_{im} + \mathbf{U}_{rl} * \mathbf{U}_{rm})); \quad \mathbf{U}_{kn} = \frac{1}{\mathbf{U}_{kk}} * (V_{kn} - (\mathbf{U}_{jm} * \mathbf{U}_{jn} + \mathbf{U}_{im} * \mathbf{U}_{in} + \mathbf{U}_{rm} * \mathbf{U}_{rn}));$$

$$\mathbf{U}_{ee} = \sqrt{V_{ee} - (\mathbf{U}_{ke}^2 + \mathbf{U}_{je}^2 + \mathbf{U}_{ie}^2 + \mathbf{U}_{re}^2)}; \quad \mathbf{U}_{el} = \frac{1}{\mathbf{U}_{ee}} * (V_{el} - (\mathbf{U}_{ke} * \mathbf{U}_{kl} + \mathbf{U}_{je} * \mathbf{U}_{jl} + \mathbf{U}_{ie} * \mathbf{U}_{il} + \mathbf{U}_{re} * \mathbf{U}_{rl}));$$

$$\mathbf{U}_{em} = \frac{1}{\mathbf{U}_{ee}} * (V_{em} - (\mathbf{U}_{kl} * \mathbf{U}_{km} + \mathbf{U}_{jl} * \mathbf{U}_{jm} + \mathbf{U}_{il} * \mathbf{U}_{im} + \mathbf{U}_{rl} * \mathbf{U}_{rm})); \quad \mathbf{U}_{en} = \frac{1}{\mathbf{U}_{ee}} * (V_{en} - (\mathbf{U}_{km} * \mathbf{U}_{kn} + \mathbf{U}_{jm} * \mathbf{U}_{jn} + \mathbf{U}_{im} * \mathbf{U}_{in} + \mathbf{U}_{rm} * \mathbf{U}_{rn}));$$

$$\mathbf{U}_{ll} = \sqrt{V_{ll} - (\mathbf{U}_{el}^2 + \mathbf{U}_{kl}^2 + \mathbf{U}_{jl}^2 + \mathbf{U}_{il}^2 + \mathbf{U}_{rl}^2)}; \quad \mathbf{U}_{lm} = \frac{1}{\mathbf{U}_{ll}} * (V_{lm} - (\mathbf{U}_{el} * \mathbf{U}_{em} + \mathbf{U}_{kl} * \mathbf{U}_{km} + \mathbf{U}_{jl} * \mathbf{U}_{jm} + \mathbf{U}_{il} * \mathbf{U}_{im} + \mathbf{U}_{rl} * \mathbf{U}_{rm}));$$

$$\mathbf{U}_{\ln} = \frac{1}{\mathbf{U}_{ll}} * (V_{\ln} - (\mathbf{U}_{em} * \mathbf{U}_{en} + \mathbf{U}_{km} * \mathbf{U}_{kn} + \mathbf{U}_{jm} * \mathbf{U}_{jn} + \mathbf{U}_{im} * \mathbf{U}_{in} + \mathbf{U}_{rm} * \mathbf{U}_{rn}));$$

$$\mathbf{U}_{mm} = \sqrt{V_{mm} - (\mathbf{U}_{lm}^2 + \mathbf{U}_{em}^2 + \mathbf{U}_{km}^2 + \mathbf{U}_{jm}^2 + \mathbf{U}_{im}^2 + \mathbf{U}_{rm}^2)}; \quad \mathbf{U}_{mn} = \frac{1}{\mathbf{U}_{mm}} * (V_{mn} - (\mathbf{U}_{lm} * \mathbf{U}_{\ln} + \mathbf{U}_{em} * \mathbf{U}_{en} + \mathbf{U}_{km} * \mathbf{U}_{kn} + \mathbf{U}_{jm} * \mathbf{U}_{jn} + \mathbf{U}_{im} * \mathbf{U}_{in} + \mathbf{U}_{rm} * \mathbf{U}_{rn}));$$

$$\mathbf{U}_{nn} = \sqrt{V_{nn} - (\mathbf{U}_{mn}^2 + \mathbf{U}_{\ln}^2 + \mathbf{U}_{en}^2 + \mathbf{U}_{kn}^2 + \mathbf{U}_{jn}^2 + \mathbf{U}_{in}^2 + \mathbf{U}_{rn}^2)}.$$

**Appendix 3** The calculation of Eq. (23) $dH_x = dr\, di\, dj\, dk\, de\, dl\, dm\, dn$

From (12) and (13), we have

$r = \rho \cos\phi_1$; $i = \rho \sin\phi_1 \cos\phi_2$; $j = \rho \sin\phi_1 \sin\phi_2 \cos\phi_3$; $k = \rho \sin\phi_1 \sin\phi_2 \sin\phi_3 \cos\phi_4$; $e = \rho \sin\phi_1 \sin\phi_2 \sin\phi_3 \sin\phi_4 \cos\phi_5$;
$l = \rho \sin\phi_1 \sin\phi_2 \sin\phi_3 \sin\phi_4 \sin\phi_5 \cos\phi_6$; $m = \rho \sin\phi_1 \sin\phi_2 \sin\phi_3 \sin\phi_4 \sin\phi_5 \sin\phi_6 \cos\phi_7$;
$n = \rho \sin\phi_1 \sin\phi_2 \sin\phi_3 \sin\phi_4 \sin\phi_5 \sin\phi_6 \sin\phi_7$

where $0 < \phi_1, \phi_2, \phi_3, \phi_4, \phi_5, \phi_6 < \pi$, $0 < \phi_7 < 2\pi$ and $\rho = \sqrt{r^2 + i^2 + j^2 + k^2 + e^2 + l^2 + m^2 + n^2}$ indicates the size of the octonion number.

Then we can evaluated $dH_x$ by Jacobian $Jf$ of $f$ defined as:

$$Jf = \frac{\partial(r,i,j,k,e,l,m,n)}{\partial(\rho,\phi_1,\phi_2,\phi_3,\phi_4,\phi_5,\phi_6,\phi_7)} = \frac{dr\, di\, dj\, dk\, de\, dl\, dm\, dn}{d\rho\, d\phi_1\, d\phi_2\, d\phi_3\, d\phi_4\, d\phi_5\, d\phi_6\, d\phi_7}$$

$$= \begin{vmatrix} \frac{dr}{d\rho} & \frac{dr}{d\phi_1} & \frac{dr}{d\phi_2} & \frac{dr}{d\phi_3} & \frac{dr}{d\phi_4} & \frac{dr}{d\phi_5} & \frac{dr}{d\phi_6} & \frac{dr}{d\phi_7} \\ \frac{di}{d\rho} & \frac{di}{d\phi_1} & \frac{di}{d\phi_2} & \frac{di}{d\phi_3} & \frac{di}{d\phi_4} & \frac{di}{d\phi_5} & \frac{di}{d\phi_6} & \frac{di}{d\phi_7} \\ \frac{dj}{d\rho} & \frac{dj}{d\phi_1} & \frac{dj}{d\phi_2} & \frac{dj}{d\phi_3} & \frac{dj}{d\phi_4} & \frac{dj}{d\phi_5} & \frac{dj}{d\phi_6} & \frac{dj}{d\phi_7} \\ \frac{dk}{d\rho} & \frac{dk}{d\phi_1} & \frac{dk}{d\phi_2} & \frac{dk}{d\phi_3} & \frac{dk}{d\phi_4} & \frac{dk}{d\phi_5} & \frac{dk}{d\phi_6} & \frac{dk}{d\phi_7} \\ \frac{de}{d\rho} & \frac{de}{d\phi_1} & \frac{de}{d\phi_2} & \frac{de}{d\phi_3} & \frac{de}{d\phi_4} & \frac{de}{d\phi_5} & \frac{de}{d\phi_6} & \frac{de}{d\phi_7} \\ \frac{dl}{d\rho} & \frac{dl}{d\phi_1} & \frac{dl}{d\phi_2} & \frac{dl}{d\phi_3} & \frac{dl}{d\phi_4} & \frac{dl}{d\phi_5} & \frac{dl}{d\phi_6} & \frac{dl}{d\phi_7} \\ \frac{dm}{d\rho} & \frac{dm}{d\phi_1} & \frac{dm}{d\phi_2} & \frac{dm}{d\phi_3} & \frac{dm}{d\phi_4} & \frac{dm}{d\phi_5} & \frac{dm}{d\phi_6} & \frac{dm}{d\phi_7} \\ \frac{dn}{d\rho} & \frac{dn}{d\phi_1} & \frac{dn}{d\phi_2} & \frac{dn}{d\phi_3} & \frac{dn}{d\phi_4} & \frac{dn}{d\phi_5} & \frac{dn}{d\phi_6} & \frac{dn}{d\phi_7} \end{vmatrix} = \begin{vmatrix} D_{r\rho} & D_{r\phi_1} & 0 & 0 & 0 & 0 & 0 & 0 \\ D_{i\rho} & D_{i\phi_1} & D_{i\phi_2} & 0 & 0 & 0 & 0 & 0 \\ D_{j\rho} & D_{j\phi_1} & D_{j\phi_2} & D_{j\phi_3} & 0 & 0 & 0 & 0 \\ D_{k\rho} & D_{k\phi_1} & D_{k\phi_2} & D_{k\phi_3} & D_{k\phi_4} & 0 & 0 & 0 \\ D_{e\rho} & D_{e\phi_1} & D_{e\phi_2} & D_{e\phi_3} & D_{e\phi_4} & D_{e\phi_5} & 0 & 0 \\ D_{l\rho} & D_{l\phi_1} & D_{l\phi_2} & D_{l\phi_3} & D_{l\phi_4} & D_{l\phi_5} & D_{l\phi_6} & 0 \\ D_{m\rho} & D_{m\phi_1} & D_{m\phi_2} & D_{m\phi_3} & D_{m\phi_4} & D_{m\phi_5} & D_{m\phi_6} & D_{m\phi_7} \\ D_{n\rho} & D_{n\phi_1} & D_{n\phi_2} & D_{n\phi_3} & D_{n\phi_4} & D_{n\phi_5} & D_{n\phi_6} & D_{n\phi_7} \end{vmatrix}$$

and the values of the entries in the determinant are shown below:

$D_{r\rho} = \cos\phi_1;\quad D_{r\phi_1} = -\rho\sin\phi_1;$

$D_{i\rho} = \sin\phi_1\cos\phi_2;\quad D_{i\phi_1} = \rho\cos\phi_1\cos\phi_2;\quad D_{i\phi_2} = -\rho\sin\phi_1\sin\phi_2;$

$D_{j\rho} = \sin\phi_1\sin\phi_2\cos\phi_3;\quad D_{j\phi_1} = \rho\cos\phi_1\sin\phi_2\cos\phi_3;\quad D_{j\phi_2} = \rho\sin\phi_1\cos\phi_2\cos\phi_3;\quad D_{j\phi_3} = -\rho\sin\phi_1\sin\phi_2\sin\phi_3;$

$D_{k\rho} = \sin\phi_1\sin\phi_2\sin\phi_3\cos\phi_4;\quad D_{k\phi_1} = \rho\cos\phi_1\sin\phi_2\sin\phi_3\cos\phi_4;\quad D_{k\phi_2} = \rho\sin\phi_1\cos\phi_2\sin\phi_3\cos\phi_4;$

$D_{k\phi_3} = \rho\sin\phi_1\sin\phi_2\cos\phi_3\cos\phi_4;\quad D_{k\phi_4} = -\rho\sin\phi_1\sin\phi_2\sin\phi_3\sin\phi_4;$

$D_{e\rho} = \sin\phi_1\sin\phi_2\sin\phi_3\sin\phi_4\cos\phi_5;\quad D_{e\phi_1} = \rho\cos\phi_1\sin\phi_2\sin\phi_3\sin\phi_4\cos\phi_5;\quad D_{e\phi_2} = \rho\sin\phi_1\cos\phi_2\sin\phi_3\sin\phi_4\cos\phi_5;$

$D_{e\phi_3} = \rho\sin\phi_1\sin\phi_2\cos\phi_3\sin\phi_4\cos\phi_5;\quad D_{e\phi_4} = \rho\sin\phi_1\sin\phi_2\sin\phi_3\cos\phi_4\cos\phi_5;\quad D_{e\phi_5} = -\rho\sin\phi_1\sin\phi_2\sin\phi_3\sin\phi_4\sin\phi_5;$

$D_{l\rho} = \sin\phi_1\sin\phi_2\sin\phi_3\sin\phi_4\sin\phi_5\cos\phi_6;\quad D_{l\phi_1} = \rho\cos\phi_1\sin\phi_2\sin\phi_3\sin\phi_4\sin\phi_5\cos\phi_6;$

$D_{l\phi_2} = \rho\sin\phi_1\cos\phi_2\sin\phi_3\sin\phi_4\sin\phi_5\cos\phi_6;\quad D_{l\phi_3} = \rho\sin\phi_1\sin\phi_2\cos\phi_3\sin\phi_4\sin\phi_5\cos\phi_6;$

$D_{l\phi_4} = \rho\sin\phi_1\sin\phi_2\sin\phi_3\cos\phi_4\sin\phi_5\cos\phi_6;\quad D_{l\phi_5} = \rho\sin\phi_1\sin\phi_2\sin\phi_3\sin\phi_4\cos\phi_5\cos\phi_6;$

$D_{l\phi_6} = -\rho\sin\phi_1\sin\phi_2\sin\phi_3\sin\phi_4\sin\phi_5\sin\phi_6;$

$D_{m\rho} = \sin\phi_1\sin\phi_2\sin\phi_3\sin\phi_4\sin\phi_5\sin\phi_6\cos\phi_7;\quad D_{m\phi_1} = \rho\cos\phi_1\sin\phi_2\sin\phi_3\sin\phi_4\sin\phi_5\sin\phi_6\cos\phi_7;$

$D_{m\phi_2} = \rho\sin\phi_1\cos\phi_2\sin\phi_3\sin\phi_4\sin\phi_5\sin\phi_6\cos\phi_7;\quad D_{m\phi_3} = \rho\sin\phi_1\sin\phi_2\cos\phi_3\sin\phi_4\sin\phi_5\sin\phi_6\cos\phi_7;$

$D_{m\phi_4} = \rho\sin\phi_1\sin\phi_2\sin\phi_3\cos\phi_4\sin\phi_5\sin\phi_6\cos\phi_7;\quad D_{m\phi_5} = \rho\sin\phi_1\sin\phi_2\sin\phi_3\sin\phi_4\cos\phi_5\sin\phi_6\cos\phi_7;$

$D_{m\phi_6} = \rho\sin\phi_1\sin\phi_2\sin\phi_3\sin\phi_4\sin\phi_5\cos\phi_6\cos\phi_7;\quad D_{m\phi_7} = -\rho\sin\phi_1\sin\phi_2\sin\phi_3\sin\phi_4\sin\phi_5\sin\phi_6\sin\phi_7;$

$D_{n\rho} = \sin\phi_1\sin\phi_2\sin\phi_3\sin\phi_4\sin\phi_5\sin\phi_6\sin\phi_7;\quad D_{n\phi_1} = \rho\cos\phi_1\sin\phi_2\sin\phi_3\sin\phi_4\sin\phi_5\sin\phi_6\sin\phi_7;$

$D_{n\phi_2} = \rho\sin\phi_1\cos\phi_2\sin\phi_3\sin\phi_4\sin\phi_5\sin\phi_6\sin\phi_7;\quad D_{n\phi_3} = \rho\sin\phi_1\sin\phi_2\cos\phi_3\sin\phi_4\sin\phi_5\sin\phi_6\sin\phi_7;$

$D_{n\phi_4} = \rho\sin\phi_1\sin\phi_2\sin\phi_3\cos\phi_4\sin\phi_5\sin\phi_6\sin\phi_7;\quad D_{n\phi_5} = \rho\sin\phi_1\sin\phi_2\sin\phi_3\sin\phi_4\cos\phi_5\sin\phi_6\sin\phi_7;$

$D_{n\phi_6} = \rho\sin\phi_1\sin\phi_2\sin\phi_3\sin\phi_4\sin\phi_5\cos\phi_6\sin\phi_7;\quad D_{n\phi_7} = \rho\sin\phi_1\sin\phi_2\sin\phi_3\sin\phi_4\sin\phi_5\sin\phi_6\cos\phi_7;$

With the lower triangular matrix rule, we obtain

$Jf = \rho^7 \sin^6\phi_1 \sin^5\phi_2 \sin^4\phi_3 \sin^3\phi_4 \sin^2\phi_5 \sin\phi_6$

Then,

$dH_x = dr\,di\,dj\,dk\,de\,dl\,dm\,dn = \rho^7 \sin^6\phi_1 \sin^5\phi_2 \sin^4\phi_3 \sin^3\phi_4 \sin^2\phi_5 \sin\phi_6\, d\rho\, d\phi_1\, d\phi_2\, d\phi_3\, d\phi_4\, d\phi_5\, d\phi_6\, d\phi_7$

**Appendix 4**  The proof of Eq. (25)

$$E[|W|^2] = \int_0^\infty x^2 f(x)dx = \int_0^\infty x^2 \frac{1}{48\sigma^8} x^7 e^{-x^2/2\sigma^2} dx$$

$$= \frac{1}{48\sigma^8} \int_0^\infty x^9 e^{-x^2/2\sigma^2} dx$$

$$= \frac{1}{48\sigma^8} \left( -x^8 \sigma^2 e^{-x^2/2\sigma^2} \Big|_0^\infty + \int_0^\infty 8x^7 \sigma^2 e^{-x^2/2\sigma^2} dx \right)$$

$$= \frac{1}{48\sigma^6} \left( -x^8 e^{-x^2/2\sigma^2} \Big|_0^\infty + \int_0^\infty 8x^7 e^{-x^2/2\sigma^2} dx \right)$$

$$= \left( -\frac{1}{48\sigma^6} x^8 e^{-x^2/2\sigma^2} \Big|_0^\infty \right) + \frac{1}{48\sigma^6} \int_0^\infty 8x^7 e^{-x^2/2\sigma^2} dx$$

$$= \frac{1}{48\sigma^6} \int_0^\infty 8x^7 e^{-x^2/2\sigma^2} dx$$

$$= \frac{1}{48\sigma^4} \int_0^\infty 8 \cdot 6 x^5 e^{-x^2/2\sigma^2} dx$$

$$= \frac{1}{48\sigma^2} \int_0^\infty 8 \cdot 6 \cdot 4 x^3 e^{-x^2/2\sigma^2} dx$$

$$= \frac{1}{48} \int_0^\infty 8 \cdot 6 \cdot 4 \cdot 2 x e^{-x^2/2\sigma^2} dx$$

$$= 8 \int_0^\infty x e^{-x^2/2\sigma^2} dx$$

$$= 8\sigma^2$$